\documentclass{article}

\usepackage{arxiv}

\usepackage[utf8]{inputenc} % allow utf-8 input
\usepackage[T1]{fontenc}    % use 8-bit T1 fonts
\usepackage{hyperref}       % hyperlinks
\usepackage{url}            % simple URL typesetting
\usepackage{booktabs}       % professional-quality tables
\usepackage{amsfonts}       % blackboard math symbols
\usepackage{nicefrac}       % compact symbols for 1/2, etc.
\usepackage{microtype}      % microtypography
\usepackage{amsmath}
\usepackage{cleveref}       % smart cross-referencing
\usepackage{lipsum}         % Can be removed after putting your text content
\usepackage{graphicx}
\usepackage[square,numbers]{natbib}
\usepackage{doi}
\usepackage{subfigure}

\title{AnoDODE: Anomaly Detection with Diffusion ODE}

\date{}

\author{ \href{}{\includegraphics[scale=0.06]{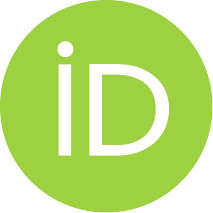}\hspace{1mm}Xianyao Hu}\\
	Department of Mathematics\\
	Zhejiang Sci-tech University\\
	\texttt{1374565032@qq.com} \\
	\And
	\href{}{\includegraphics[scale=0.06]{orcid.pdf}\hspace{1mm}Congming Jin} \\
	Department of Mathematics\\
	Zhejiang Sci-tech University\\
	\texttt{jincm@zstu.edu.cn} \\
}

\renewcommand{\headeright}{}
\renewcommand{\undertitle}{}
\hypersetup{
pdftitle={A template for the arxiv style},
pdfsubject={q-bio.NC, q-bio.QM},
pdfauthor={David S.~Hippocampus, Elias D.~Striatum},
pdfkeywords={First keyword, Second keyword, More},
}

\begin{document}
\maketitle

\begin{abstract}
	
	Anomaly detection is the process of identifying atypical data samples that significantly deviate from the majority of the dataset. In the realm of clinical screening and diagnosis, detecting abnormalities in medical images holds great importance. Typically, clinical practice provides access to a vast collection of normal images, while abnormal images are relatively scarce.	
	We hypothesize that abnormal images and their associated features tend to manifest in low-density regions of the data distribution. Following this assumption, we turn to diffusion ODEs for unsupervised anomaly detection, given their tractability and superior performance in density estimation tasks.
	More precisely, we propose a new anomaly detection method based on diffusion ODEs by estimating the density of features extracted from multi-scale medical images.
	Our anomaly scoring mechanism depends on computing the negative log-likelihood of features extracted from medical images at different scales, quantified in bits per dimension.
	Furthermore, we propose a reconstruction-based anomaly localization suitable for our method. Our proposed method not only identifie anomalies but also provides interpretability at both the image and pixel levels.
	Through experiments on the BraTS2021 medical dataset, our proposed method outperforms existing methods. These results confirm the effectiveness and robustness of our method.
	
\end{abstract}

% keywords can be removed

\section{INTRODUCTION}

Anomaly detection is an important technique used to identify patterns or instances that deviate significantly from majority of the dataset. It plays an important role in a wide range of real-world applications, including but not limited to video surveillance, manufacturing inspection, rare disease detection and autonomous driving. In the current field of anomaly detection research, unsupervised paradigms dominate, where models are trained only on normal samples. This approach has significant advantages in labeling efficiency, making it particularly valuable in the field of computational medical image analysis.

In medical anomaly detection, anomalies may include abnormal structures, lesions, or patterns that may indicate the presence of disease. In the real world, it is often easier to collect normal data than to obtain abnormal samples, especially those related to rare diseases. Unsupervised anomaly detection reduces the reliance on label-balanced medical datasets. Hence, it has far-reaching significance in medical diagnosis. Unsupervised image anomaly detection methods mostly use image features extracted by pre-trained encoders. Density estimation of image features is a paradigm for unsupervised image anomaly detection. It assumes that abnormal images and their features appear in low-density areas, and estimates the negative log-likelihood of the test dataset as the basis for anomaly detection. Choosing a better density estimator often leads to better anomaly detection performance.

Normalizing flow \cite{dinh2014nice, dinh2016density,rezende2015variational,kingma2018glow} is a popular method that transform the observed data distribution to a tractable distribution (e.g., a standard normal distribution) by a sequence of invertible and differentiable mappings. Chen et al. \cite{chen2018neural} introduced a continuous-time analog of normalizing flows, defining the mapping from data to latent variables using ordinary differential equations (ODEs). The log-likelihood can be computed using trace operations.  Grathwohl et al. \cite{grathwohl2018ffjord} used Hutchinson’s trace estimator to give a scalable unbiased estimate of the log-likelihood and achieved more accurate density estimation performance.
Song et al. \cite{song2020score} proposed diffuion ODE and experimentally proved that it has better density estimation performance than normalizing flow.
Diffusion ODE describes the evolution of the marginal probability density of a diffusion process.
Since diffusion ODE is a deterministic process that converts the distribution of data into a simple distribution, similar to the continuous normalizing flow, it can estimate the likelihood using the Hutchinson trace estimator.

In this paper, we explore diffusion ODE which is the latest density estimator for medical image anomaly detection. Like most methods, we estimate the density of image features extracted by a pre-trained encoder and consider anomalies to occur in low-density regions. In addition, in medical practice, misdiagnosis can have serious consequences, so the robustness of anomaly detection models is particularly important. We adopt a multi-scale strategy to increase the robustness of our method. We demonstrate the effectiveness and robustness of our method through experiments on the BraTS2021 medical dataset.

\section{RELATED WORK}

In the field of unsupervised anomaly detection, existing deep learning methods mainly include five types: reconstruction based methods, knowledge distillation based methods, memory bank based methods, one class classification based methods, likelihood based methods. Most methods are designed to solve industrial anomaly detection problems, and they are still very powerful when applied to medical image anomaly detection with similar data structures.

Reconstruction based anomaly detection trains a generative model, such as autoencoder (AE) \cite{lecun1989generalization} and generative adversarial network (GAN) \cite{goodfellow2014generative}, to learn normal data patterns and spots anomalies by detecting high reconstruction errors, which signal deviations from the normal data. AE-based anomaly detection methods initially took the spotlight due to their role in compressing information within the encoder-decoder architecture, making them popular choices.
Later, GAN-based anomaly detection methods became famous for their ability to generate high-quality samples. AnoGAN\cite{schlegl2017unsupervised} is the earliest use of GAN for anomaly detection by comparing original data with synthetic data. GANomaly\cite{akcay2019ganomaly} combined GAN with an encoder-decoder architecture to incorporate distances in latent feature space to distinguish anomalous data. F-AnoGAN\cite{schlegl2019f} combined GAN with simultaneous encoder guidance in image and latent space to improve the accuracy and efficiency of identifying anomalies in image datasets.
Recently, there is a trend to utilize diffusion models for anomaly detection. When using diffusion model for anomaly detection, the reconstructed image is obtained by injecting noise and denoising and large reconstruction errors indicate the presence of anomalies. AnoDDPM\cite{wyatt2022anoddpm} utilized denoising diffusion probabilistic models (DDPMs) with multi-scale simplex noise to detecte anomalies.  DDAD\cite{mousakhan2023anomaly} built conditioned DDPMs that are conditioned on the input image to achieve highly accurate recovery while preserving the image structure.

The remaining four types of methods typically model abstract representations which map data into an embedding space using a pre-trained network or a task-specific model and achieve superior performance in image anomaly detection.
One-class classification anomaly detection methods, like the one-class support vector machine \cite{scholkopf2001estimating}, support vector data description \cite{tax2004support}, DeepSVDD \cite{ruff2018deep}, and PatchSVDD \cite{yi2020patch}, utilized normal data to create compact representations, such as hyperplanes or hyperspheres, in high-dimensional feature spaces and classified any datas falling outside these representations as anomalies. CutPaste\cite{li2021cutpaste} proposed an effective technique for extending one-class classification by generating synthetic abnormalities, which are used as negative samples during model training.
In knowledge distillation based methods, the Teacher-Student architecture distills knowledge from a teacher model to a student network, using representation differences to detect anomalies. MKD\cite{salehi2021multiresolution} leveraged multi-scale feature discrepancy between the teacher-student pair for anomaly detection. RD4AD\cite{deng2022anomaly} proposed a novel Teacher-Student model with a "reverse distillation" approach and a trainable one-class bottleneck embedding, significantly improving performance.
Memory Bank based method uses a memory bank to store and compare feature representations of normal data during training, enabling the detection of anomalies in new data based on deviations from the stored representations.
PaDiM\cite{defard2021padim} combined patch distribution modeling with pretrained neural networks.
PatchCore\cite{roth2022towards} utilized core set sampling to build a memory bank and nearest neighbor search for inference.

Likelihood based methods rely on the assumption that abnormal data and its features are concentrated in low-density areas, and it explicitly models the distribution of the data or its features. Many methods exploit the power of the normalizing flow, a deep learning density estimation that has gained popularity in recent years. DifferNet \cite{rudolph2021same} processed input images at multi-scale using a feature extractor, and then use a normalizing flow to transform the concatenated feature outputs into a normal distribution via maximum likelihood training. The anomaly score is defined as the mean of the negative log-likelihood. CS-Flow\cite{rudolph2022fully} processed the multi-scale feature maps jointly, using a fully convolutional normalizing flow with cross-connections between scales. AE-FLOW \cite{zhao2023ae} constructed a loss function and anomaly score function by an autoencoder with the normalizing flow. Although normalizing flow-based anomaly detection methods achieve excellent performance, we believe that the performance will be further enhanced with the emergence of better density estimators.

\section{PRELIMINARIES}

\begin{figure}
	\centering
	\includegraphics[width=1\linewidth]{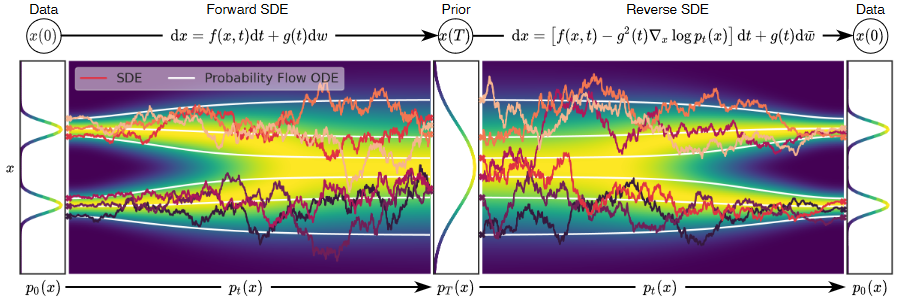}
	\caption{Overview of score-based generative modeling through SDEs.}
	\label{score}
\end{figure}

DDPMs \cite{ho2020denoising,sohl2015deep} have been shown to achieve state-of-the-art performance on various image generation tasks and are known for their ability to generate high-quality and diverse samples. DDPMs consist of a forward diffusion process gradually corrupting data from a target distribution into a normal distribution, and a learned reverse process that generates samples by turning noise into samples. Song et al. \cite{song2020score} extended this approach to a continuous formulation and construct a diffusion process $\mathbf{x}(t)$ with a continuous time variable $t\in\left[0,T\right]$.
The probability density of $\mathbf{x}(t)$ is denoted by $p_t(\mathbf{x})$, i.e., $\mathbf{x}(t)\sim p_{t}(\mathbf{x})$ and $\mathbf{x}(0)\sim p_{0}(\mathbf{x})$, for which we have a dataset of i.i.d. samples, and $\mathbf{x}(T)\sim p_{T}(\mathbf{x})$ , for which we have a tractable form to generate samples efficiently. In other words, $p_0$ is the data distribution and $p_T$ is the prior distribution. Usually, $p_T$ is an unstructured prior distribution that has no knowledge of $p_0$. For instance, it could be a Gaussian distribution with fixed mean and variance.
The overview of this framework named score-based generative model can be seen in Figure \ref{score} which comes from \cite{song2020score}.
The forward diffusion process can be redefined as a process given by a stochastic differential equation (SDE):
\begin{equation}\label{f_sde}
	\mathrm{d} \mathbf{x}=\mathbf{f}(\mathbf{x}, t) \mathrm{d} t+g(t) \mathrm{d} \mathbf{w},
\end{equation}
where $\mathbf{w}$ is the standard Wiener process, $f(\cdot,t)$ is a vectorvalued function called the drift coefficient of $\mathbf{x}(t)$, and $g(t)$ is a scalar function known as the diffusion coefficient of $\mathbf{x}(t)$. The SDE (\ref{f_sde}) has a unique strong solution as long as $f(\cdot,t)$ and $g(t)$ are globally Lipschitz in both state and time \cite{oksendal2003stochastic}.
The transition kernel from $\mathbf{x}(s)$ to $\mathbf{x}(t)$ is denoted by $p_{st}(\mathbf{x}(t) | \mathbf{x}(s))$, where $0 \le s < t \le T$.

Anderson \cite{anderson1982reverse} stated that the reverse of a diffusion process is also a diffusion process, running backwards in time and given by the reverse-time SDE:
\begin{equation}\label{r_sde}
	\mathrm{d} \mathbf{x}=\left[\mathbf{f}(\mathbf{x}, t)-g(t)^{2} \nabla_{\mathbf{x}} \log p_{t}(\mathbf{x})\right] \mathrm{d} t+g(t) \mathrm{d} \bar{\mathbf{w}},
\end{equation}
where $\bar{\mathbf{w}}$ is a standard Wiener process when time flows backwards from $T$ to $0$, and $\mathrm{d} t$ is an infinitesimal negative timestep. To estimate $\nabla_{\mathbf{x}} \log p_{t}(\mathbf{x})$, which is the score of each marginal distribution $ p_{t}(\mathbf{x})$ for all $t$, we can train a time-dependent score-based model $\mathbf{s}_{\boldsymbol{\theta}}(\mathbf{x}(t), t)$ to approach it via the continuous objective:
\begin{equation}\label{objective}
	\min_\theta\mathbb{E}_{t\sim\mathcal{U}(0,T)}[\lambda(t)\mathbb{E}_{\mathbf{x}(0)\sim p_0(\mathbf{x})}\mathbb{E}_{\mathbf{x}(t)\sim p_{0t}(\mathbf{x}(t)\mid\mathbf{x}(0))}[\|s_\theta(\mathbf{x}(t),t)-\nabla_{\mathbf{x}(t)}\log p_{0t}(\mathbf{x}(t)|\mathbf{x}(0))\|_2^2]],
\end{equation}
where $\mathcal{U}(0,T)$ is a uniform distribution over $[0,T]$, and $\lambda(t):[0, T] \rightarrow \mathbb{R}_{>0}$ is a positive weighting function, often neglected in practice. Then, we can obtain samples $\mathbf{x}(0)\sim p_{0}$ by starting from samples of $\mathbf{x}(T)\sim p_{T}$ and reversing the reverse-time SDE Eq.(\ref{r_sde}).

Song et al. \cite{song2020score} showed that the reverse diffusion process of the SDE (\ref{r_sde}) can be modeled as a deterministic process as the marginal probabilities can be modeled deterministically in terms of the score function. As a result, the problem simplifies to an ODE called probability flow ODE:
\begin{equation}\label{score ode}
	\mathrm{d} \mathbf{x}=\left[\mathbf{f}(\mathbf{x}, t)-\frac{1}{2} g(t)^{2} \nabla_{\mathbf{x}} \log p_{t}(\mathbf{x})\right] \mathrm{d} t.
\end{equation}
Similarly, the score network $\mathbf{s}_{\boldsymbol{\theta}}(\mathbf{x}(t), t)$ can be trained to fit the true score function $\nabla_{\mathbf{x}} \log p_{t}(\mathbf{x})$ for all $t$, resulting in the diffusion ODE:
\begin{equation}\label{diffusion ode}
	\mathrm{d} \mathbf{x}=\left[\mathbf{f}(\mathbf{x}, t)-\frac{1}{2} g(t)^{2} \mathbf{s}_{\boldsymbol{\theta}}(\mathbf{x}(t), t)\right] \mathrm{d} t.
\end{equation}
Diffusion ODE is a special case of the continuous normalizing flow \cite{chen2018neural}, thus can compute the exact likelihood on any input data. Experiments\cite{song2020score, song2021maximum} demonstrated the excellent density estimation performance of diffusion ODE.

\section{METHOD}
\begin{figure}
	\centering
	\includegraphics[width=0.8\linewidth]{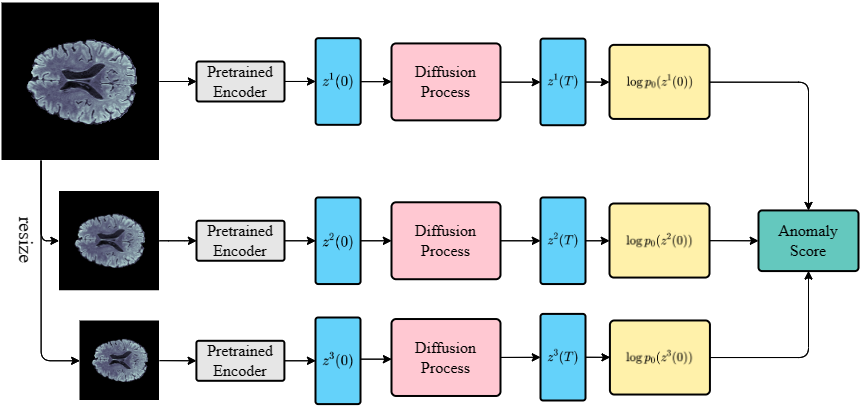}
	\caption{The overview of the AnoDODE model.
		We model the features $\mathbf{z}^{k}(0), k=1,2,3$ which are extracted from medical images $\mathbf{x}^{k}, k=1,2,3$ at different scales using pre-trained encoders. The first step involves a diffusion process that injects noise into these features and subsequently learns the inverse diffusion process via a score network architecture like U-net. Following, we calculate the log-likelihood of the features using diffusion ODE.		
		Finally, we compute the anomaly score which is the average of the likelihoods of the features at different scales.}
	\label{model}
\end{figure}

We propose a new anomaly detection method that utilizes diffusion ODE for density estimation on feature vectors extracted from medical images at different scales using pre-trained feature extractors. The underlying intuition is that feature vectors from normal images tend to map to the high-density region of the standard Gaussian distribution as the diffusion process is trained on normal data. Conversely, feature vectors from abnormal images tend to map to the distribution's tail, resulting in lower likelihood values.
Utilizing a multi-scale modeling approach allows us to capture both global and local information within medical images across different scales, enhancing the model's robustness. The anomaly score is computed as the negative log-likelihoods measured in bits per dimension for features extracted from multi-scale images.
Our proposed pipeline comprises two crucial components: the encoder block and the diffusion process block. For a overview of the anomaly detection with diffusion ODE (AnoDODE), please refer to Figure \ref{model}.
Additionally, we propose a reconstruction-based anomaly localization method tailored to our model. This involves training a decoder for the dataset, and you can find further details in Section \ref{localization}.
In the following subsections, we will provide a detailed explanation of the diffusion blocks, anomaly score and localization.

\subsection{Diffusion Process}

The diffusion process is to continuously inject noise at continuous time $t$ in features $\textbf{z}$ extracted from multi-scale medical images to obtain a simple distribution. We set this simple distribution to be a standard Gaussian distribution with a mean of $0$ and a variance of $1$. The noise injection method of our model adopts the Variance Preserving (VP) SDE proposed by Song et al.\cite{song2020score}, which is a continuous form of the discrete injection method in DDPM. The forward diffusion process of VPSDE is \begin{equation}\label{f_vpsde}
	\mathrm{d} \mathbf{z}=-\frac{1}{2} \beta(t) \mathbf{z} \mathrm{d} t+\sqrt{\beta(t)} \mathrm{d} \mathbf{w},
\end{equation}
where $\beta(t)=\bar{\beta}_{\min}+t(\bar{\beta}_{\max}-\bar{\beta}_{\min})$ with $\bar{\beta}_{\mathrm{min}}=0.1$ and $\bar{\beta}_{\mathrm{max}}=20$ for $t\in[0,1]$, $\mathbf{w}$ is the standard Wiener process.
The corresponding reverse diffusion process of VPSDE is
\begin{equation}\label{r_vpsde}
\mathrm{d} \mathbf{z}=\left[-\frac{1}{2} \beta(t) \mathbf{z} -\beta(t) \nabla_{\mathbf{z}} \log p_{t}(\mathbf{z})\right] \mathrm{d} t+\sqrt{\beta(t)} \mathrm{d} \bar{\mathbf{w}},
\end{equation}
where $\bar{\mathbf{w}}$ is a standard Wiener process when time flows backwards from $1$ to $0$.
The only unknown term is the score function $\nabla_{\mathbf{z}} \log p_{t}(\mathbf{z})$. We fit $\nabla_{\mathbf{z}} \log p_{t}(\mathbf{z})$ by the output  $\mathbf{s}_{\boldsymbol{\theta}}(\mathbf{z}(t), t)$ of a U-Net in experiments. Since VPSDE has affine drift coefficients, the corresponding perturbation kernels $p_{0 t}(\mathbf{z}(t)| \mathbf{z}(0))$ are all Gaussian and can be computed in \cite{sarkka2019applied}:
\begin{equation}\label{p0t}
	p_{0t}(\mathbf{z}(t)|\mathbf{z}(0))=\mathcal{N}\left(\mathbf{z}(t);\mathbf{z}(0)e^{-\frac{1}{2}\int_0^t\beta(s)\mathrm{d}s},\mathbf{I}-\mathbf{I}e^{-\int_0^t\beta(s)\mathrm{d}s}\right).
\end{equation}
During training, $\mathbf{s}_{\boldsymbol{\theta}}(\mathbf{z}(t), t)$ can be efficiently trained with the objective in Eq.(\ref{objective}) that measures the discrepancy between the generated samples and the real data samples.

\subsection{Anomaly Score}
The diffusion ODE in Eq.(\ref{diffusion ode}) has the following form:
\begin{equation}\label{s ode}
	\mathrm{d}\mathbf{z}=\underbrace{\left\{-\frac{1}{2} \beta(t) \mathbf{z}-\frac{1}{2} \beta(t) s_\theta(\mathbf{z}(t),t)\right\}}_{=:\tilde{\mathbf{f}}_{\boldsymbol{\theta}}(\mathbf{z},t)}\mathrm{d}t.
\end{equation}
With the instantaneous change of variables formula \cite{chen2018neural}, we can compute the log-likelihood of $p_0(\mathbf{z})$ using
\begin{equation}\label{change of variables}
	\log p_0(\mathbf{z}(0))=\log p_T(\mathbf{z}(T))+\int_0^T\nabla\cdot\tilde{\mathbf{f}}_{\boldsymbol{\theta}}(\mathbf{z}(t),t)\mathrm{d}t,
\end{equation}
where the feature $\mathbf{z}(t)$ as a function of $t$ can be obtained by solving the diffusion ODE in Eq.(\ref{s ode}) and $\nabla\tilde{\mathbf{f}}_{\boldsymbol{\theta}}$ denotes the Jacobian of $\tilde{\mathbf{f}}_{\boldsymbol{\theta}}(\cdot,t)$. In many cases computing $\nabla\cdot\tilde{\mathbf{f}}_{\boldsymbol{\theta}}(\mathbf{z},t)$ is expensive, so we follow Grathwohl et al.\cite{grathwohl2018ffjord} to estimate it with the Skilling-Hutchinson trace estimator \cite{hutchinson1989stochastic, skilling1989eigenvalues}. In particular, we have
\begin{equation}\label{trace estimator}
   \nabla\cdot\tilde{\mathbf{f}}_{\boldsymbol{\theta}}(\mathbf{z},t)=\mathbb{E}_{p(\boldsymbol{\epsilon})}[\boldsymbol{\epsilon}^\mathsf{T}\nabla\tilde{\mathbf{f}}_{\boldsymbol{\theta}}(\mathbf{z},t)\boldsymbol{\epsilon}],
\end{equation}
where the random variable $\boldsymbol{\epsilon}$ satisfies $\mathbb{E}_{p(\boldsymbol{\epsilon})}[\boldsymbol{\epsilon}]=\mathbf{0}$ and $\mathrm{Cov}_{p(\boldsymbol{\epsilon})}[\boldsymbol{\epsilon}]=\mathbf{I}$. The vector-Jacobian product $\boldsymbol{\epsilon}^\mathsf{T}\nabla\tilde{\mathbf{f}}_{\boldsymbol{\theta}}(\mathbf{z},t)$ can be efficiently computed using reverse-mode automatic differentiation, at approximately the same cost as evaluating $\tilde{\mathbf{f}}_{\boldsymbol{\theta}}(\mathbf{z},t)$. As a result we can sample $\boldsymbol{\epsilon}\sim p(\boldsymbol{\epsilon})$ and then compute an efficient unbiased estimate to $\nabla\cdot\tilde{\mathbf{f}}_{\boldsymbol{\theta}}(\mathbf{z},t)$ using $\boldsymbol{\epsilon}^{\mathsf{T}}\nabla\tilde{\mathbf{f}}_{\boldsymbol{\theta}}(\mathbf{z},t)\boldsymbol{\epsilon}$. We can compute the log-likelihood of the features by substituting the Skilling-Hutchinson estimator Eq.(\ref{trace estimator}) into Eq.(\ref{change of variables}), since this estimator is unbiased.

Our anomaly scoring mechanism depends on computing the bits per dimension (bpd) of features extracted from medical images at different scales. In order to calculate the bpd score, we can compute the negative log-likelihood and change the log base (as bits are binary while negative log-likelihood is usually exponential):
\begin{equation}\label{bpd}
	f_{\mathrm{bpd}}(\mathbf{z}(0))=-\log p_0(\mathbf{z}(0))/\left(\log 2\cdot\prod_{i=1}^3 d_i\right)
\end{equation}
where $d_1,d_2,d_3$ represent the height, width and number of channels of the image features respectively.
The anomaly score is defined as the average value of the bpd of the features at different scales:
\begin{equation}\label{anomaly score}
	S = \left(f_{\mathrm{bpd}}(\mathbf{z}^{1}(0)) + f_{\mathrm{bpd}}(\mathbf{z}^{2}(0)) +\cdots + f_{\mathrm{bpd}}(\mathbf{z}^{n}(0))\right)/n
\end{equation}
where $\mathbf{z}^{1}(0),\mathbf{z}^{2}(0),\dots, \mathbf{z}^{n}(0)$ represent the features of images on different scales extracted by the pre-trained feature extractor.

\subsection{Localization}
\label{localization}

\begin{figure}
	\centering
	\includegraphics[width=0.8\linewidth]{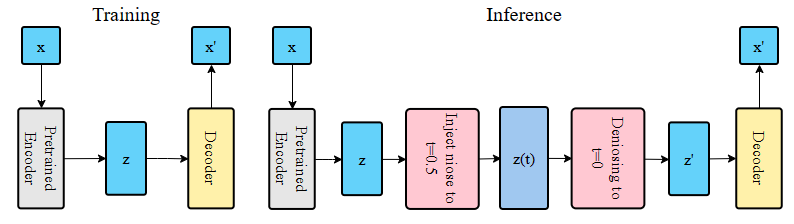}
	\caption{The pipeline of our anomaly localization method. In the training phase, we train a decoder to recover image features extracted by a pre-trained encoder. In the inference phase, we inject noise into the features in continuous time and use sampling method to reconstruct the features. Then, our trained decoder restores the reconstructed features to generate images. The difference between the original image and the reconstructed image helps us locate anomalies.}
	\label{decoder}
\end{figure}
We propose a reconstruction-based anomaly localization tailored to our method. Given our training of the score network $s_\theta(\mathbf{z}(t),t)$, we can achieve image features reconstruction by injecting noise and subsequently removing noise. Using normal images, we further train a decoder that is able to recover these reconstructed features to generate images.
Since our diffusion process and decoder are trained with normal images, abnormal images will fall into the distribution of normal images after reconstruction, resulting in reconstructed images that are significantly different from the original abnormal images.
To assess the quality of these reconstructions, we compute the squared error $(\mathbf{x}-\mathbf{x}')^2$ between the reconstructed image $\mathbf{x}'$ and the original image $\mathbf{x}$. This squared error can be visually depicted as a heatmap. The squared error $(\mathbf{x}-\mathbf{x}')^2$ provides valuable insights into the localization of anomalies in the images on pixel-level.
For a comprehensive overview of the model pipeline, please refer to Figure \ref{decoder}. Experimental results related to anomaly localization are presented in the section \ref{DETECTION}.

We use a Predictor-Corrector (PC) sampler proposed by Song et al.\cite{song2020score} as our denoising method. In the PC sampler, the sampling process involves two steps: the predictor step and the corrector step.
In the prediction step, we apply Euler-Maruyama formula:
\begin{equation}\label{E-M}
	\mathbf{z}_{t}=\left(1+\frac{1}{2}\beta_{t+1}\right)\mathbf{z}_{t+1}+\beta_{t+1}\mathbf{s}_{\boldsymbol{\theta}^{*}}(\mathbf{z}_{t+1},t+1)+\sqrt{\beta_{t+1}}\boldsymbol{\epsilon}_{t+1},
\end{equation}
where $\mathbf{s}_{\boldsymbol{\theta}^{*}}(\mathbf{z}_{t+1},t+1))$ is our trained score network and $\boldsymbol{\epsilon}_{t+1}\sim\mathcal{N}(\mathbf{0},\mathbf{I})$.
It is a general-purpose numerical method for solving the reverse-time SDE (\ref{r_vpsde}) for sample generation.
In the corrector step, since we have a score-based model $\mathbf{s}_{\boldsymbol{\theta}*}(\mathbf{z},t)\approx\nabla_{\mathbf{z}}\operatorname{log}p_{t}(\mathbf{z})$, we can employ Langevin MCMC \cite{parisi1981correlation, grenander1994representations, song2019generative} to sample from $p_t$ directly:
\begin{equation}\label{correct}
	\mathbf{z}'_{t}=\mathbf{z}_{t}+h_{t}\mathbf{s}_{\boldsymbol{\theta}*}(\mathbf{z}_{t},t)+\sqrt{2h_{t}}\boldsymbol{\epsilon}_t,
\end{equation}
where $\mathbf{z}'_{t}$ is the corrected  $\mathbf{z}_{t}$, $h_t$ is the step size, defined as $2(r\sqrt{d}/\left\|\mathbf{s}_{\boldsymbol{\theta}*}(\mathbf{z}_{t},t)\right\|_2)^2$ where $d$ is the dimensionality of $\mathbf{z}_{t}$ and $r$ is the signal-to-noise ratio, and $\boldsymbol{\epsilon}_{t}\sim\mathcal{N}(0,\mathbf{I})$.
This step can correct the solution of a numerical SDE solver.

It's important to note that you have the flexibility to choose the end time for injecting noise, but this decision requires a careful balance. Injecting noise too close to the end of the process can risk damaging the structural information within the image. Conversely, selecting an excessively short duration may not allow the generation of features within the normal image feature distribution.

\section{EXPERIMENTS}
\subsection{DATASETS}

BraTS2021\cite{baid2021rsna}, short for Brain Tumor Segmentation Challenge 2021, represents an annual competition and benchmark in the field of medical image analysis. BraTS2021 dataset consists of 3D multimodal magnetic resonance imaging (MRI) scans. We use 2D axial slices extracted from these 3D MRI scans given in \cite{bao2023bmad}. The dataset only contains slices within the depth range between 60 and 100 for data quality.
The training dataset contains 7,500 healthy slices, while the test dataset contains 3,075 tumor or abnormal slices and 640 healthy slices.

\subsection{IMPLEMENTATION DETAILS}
\label{IMPLEMENTATION DETAILS}

We utilize the output of an EfficientNet-B5 \cite{tan2019efficientnet} of layer 36 as the feature extractor for all experiments as it provides feature maps having a good balance between level of feature semantic and spatial resolution. The parameters of EfficientNet-B5 are fixed during the experiments which was pre-trained on ImageNet\cite{deng2009imagenet}.
In the case of BraTS2021, we utilize features at four scales ($n=4$) with input image sizes of $256\times256$, $192\times192$, $128\times128$ and $64\times64$ pixels. This results in feature maps with spatial dimensions of $8\times8$, $6\times6$, $4\times4$ and $2\times2$, each containing 304 channels.
In our experiments, we adopt the same UNet architecture from Dhariwal and Nichol\cite{dhariwal2021diffusion} for approximating $\nabla_{\mathbf{x}} \log p_{t}(\mathbf{x})$ and the same decoder architecture from \cite{rombach2022high}. These architectures draw inspiration from PixelCNN\cite{salimans2017pixelcnn++} and Wide ResNet\cite{zagoruyko2016wide}, with the addition of transformer sinusoidal positional embedding \cite{vaswani2017attention}.
In order to calculate the likelihood of image features, we employ the RK45 ODE solver\cite{dormand1980family} from $\mathsf{scipy.integrate.solve\_ivp}$ to solve the ODE (\ref{s ode}).
In the PC sampler, we alternate between the predictor and corrector. We use the reverse diffusion SDE solver Euler-Maruyama formula as the predictor and employ Langevin MCMC with a signal-to-noise ratio of $r=0.16$ as the corrector. The number of sampling steps is set to $N=500$.
For optimization, we use Adam \cite{kingma2014adam} with a learning rate of $10^{-4}$ in all cases.

\subsection{RESULTS}

% \usepackage{graphicx}
\begin{table}[]
	\centering
	\caption{Anomaly detection performance comparison for different methods on BraTS2021.}
	\renewcommand\arraystretch{1.8}
	\resizebox{\textwidth}{!}{%
		\begin{tabular}{cccccccc}
			\hline
			& Autoencoder & f-AnoGAN\cite{schlegl2019f} & Cutpaste\cite{li2021cutpaste} & MKD\cite{salehi2021multiresolution}   & RD4AD\cite{deng2022anomaly} & CS-Flow\cite{rudolph2022fully} & AnoDODE(ours) \\ \hline
			AUROC(\%) & 65.54       & 80.96    & 79.17    & 81.26 & 89.34 & 90.88   & \textbf{92.19}        \\
			F1-score(\%)    & 90.64       & 91.63    & 91.06    & 92.02 & 92.64 & 92.60   & \textbf{93.40}         \\
			ACC(\%)   & 82.93       & 85.06    & 83.93    & 85.84 & 87.24 & 87.51   & \textbf{89.02}         \\ \hline
		\end{tabular}%
	}
\label{result}
\end{table}

\begin{figure}[t]
	\centering
	\subfigure{
		\label{Fig.sub.1}
		\includegraphics[width=0.5\linewidth]{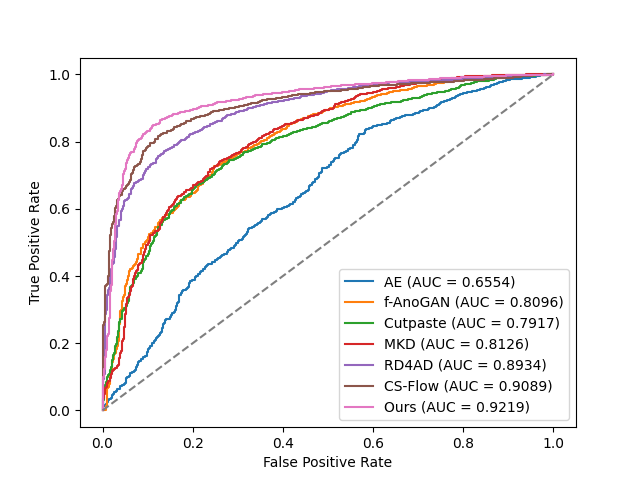}}\subfigure{
		\label{Fig.sub.2}
		\includegraphics[width=0.5\linewidth]{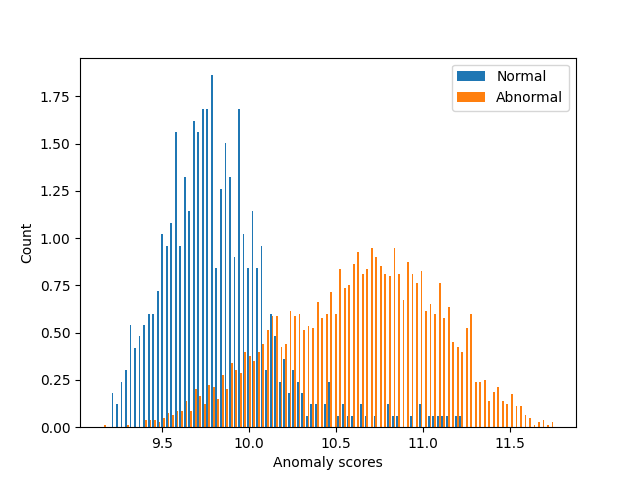}}
	\caption{\textbf{Left:} ROC-Curves of the methods on BraTS2021. Our method achieves the best results among all methods. \textbf{Right:} Distribution curves of the anomaly score of the anomaly and normal data. We can see two distributions have different peaks and the ratio of overlapping areas is relatively small. }
	\label{auc_hist}
\end{figure}

The numerical results of our method are compared with the results of six deep learning-based methods: two reconstruction based methods (Autoencoder and f-AnoGAN), two knowledge distillation based methods (MKD and RD4AD), one class classification based method (Cutpaste) and one likelihood based method (CS-Flow).
Several metrics are employed to assess the performance, including the Area Under the Receiver Operating Characteristic curve (AUROC), F1-score, and average classification accuracy (ACC). The threshold used for ACC is determined based on the optimal F1-score.
All results reported in this paper based on the best performance achieved on the test dataset. For a detailed comparison among the various methods on BraTS2021, please refer to Table \ref{result}. Additionally, Figure \ref{auc_hist} illustrates the corresponding ROC curves of all models and distribution curves of our method on BraTS2021.

On the BraTS2021 dataset, our AnoDODE method demonstrates state-of-the-art performance on all metrics. It is worth noting that the AE method shows the worst performance, but it may produce seemingly good F1-scores and ACC values due to the imbalanced distribution of positive and negative samples in the test dataset. The performance of f-AnoGAN was also unsatisfactory, possibly due to its reliance on GANs, which are known to bring training instability. Cutpaste does not achieve good performance, probably because the way it synthesizes anomalies is not suitable for the BraTS2021 dataset.
MKD and RD4AD are anomaly detection methods based on knowledge distillation, which require a validation set to prevent overfitting. RD4AD performs well on the BraTS2021 data set while MKD performs poorly.
Comparatively, CS-Flow and our method have the same principle, and both use density estimation of image features for anomaly detection. However, CS-Flow relies on normalizing flow, while our method is based on diffusion ODE that perform better in density estimation. As expected, our method outperforms CS-Flow on all metrics.

\subsection{DETECTION}
\label{DETECTION}

\begin{figure}
	\centering
	\includegraphics[width=0.8\linewidth]{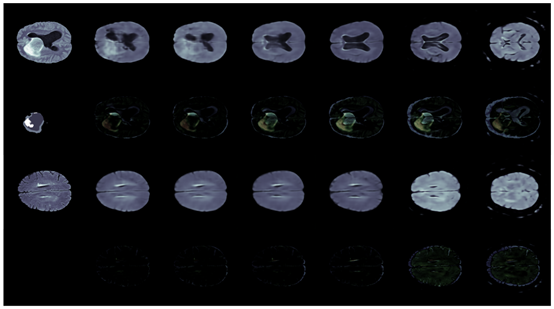}
	\caption{The localization comparison of normal and abnormal samples from the BraTS2021 dataset. The first row displays an original anomaly image, followed by the image reconstructed solely by the autoencoder, and subsequently, the images reconstructed with diffusion times 0.1, 0.3, 0.5, 0.7, and 0.9. The second row presents the ground truth alongside the corresponding heatmap, which represents the squared error between the reconstructed image and the original image.
	The third and fourth rows showcase a normal image, its reconstruction, and the corresponding heatmap.}
	\label{detection}
\end{figure}

In Section \ref{localization}, we propose an anomaly localization method that is suitable for our method. By additionally training a decoder, we can obtain image-level reconstruction. The experimental results based on the BraTS2021 dataset are shown in Figure \ref{detection}. Our results show that for abnormal images, it is difficult to detect abnormalities in reconstructions without diffusion processes (only AE processes) and reconstructions with short diffusion times. In contrast, reconstructions with long diffusion times struggle to faithfully reproduce the complete structure of the original image. We consider the optimal diffusion time to be at around 0.5. This choice strikes a balance between preserving structural information and ensuring that the generated features conform to the distribution of features found in normal images. Furthermore, when comparing the normal image with its reconstructed image, the reconstructed image is very similar to the input image and the reconstruction error is negligible. This difference helps distinguish anomalies and enhance the visual interpretability of the detection model.
\subsection{ABLATION STUDY}

% \usepackage{graphicx}
\begin{table}[]
	\centering
	\renewcommand\arraystretch{1.8}
	\caption{Ablation study on BraTS2021 with varying strategies of multi-scale. The first four columns are AnoDODE on single-scale, and the last two columns are AnoDODE on multi-scale. $\text{AnoDODE}_{128,192}$ denotes anomaly detection with diffusion ODE on $128\times128$ scale and $192\times 192$ scale and $\text{AnoDODE}_{64,128,192,256}$ is based on four scales.}
	\resizebox{\textwidth}{!}{%
		\begin{tabular}{ccccccc}
			\hline
			& $\text{AnoDODE}_{256\times 256}$ & $\text{AnoDODE}_{192\times 192}$ & $\text{AnoDODE}_{128\times 128}$ & $\text{AnoDODE}_{64\times 64}$ & $\text{AnoDODE}_{64,128,192,256}$ & $\text{AnoDODE}_{128,192}$ \\ \hline
			AUROC(\%) & 90.62                  & 91.70                  & 91.81                  & 86.95                & 91.90                      & $\textbf{92.19}$               \\
			F1-score(\%)    & 90.31                  & 93.38                  & 93.16                  & 91.99                & $\textbf{93.47}$                      & 93.40               \\
			ACC(\%)   & 88.53                  & 88.80                  & 88.45                  & 85.73                & 88.94                      & $\textbf{89.01}$              \\ \hline
		\end{tabular}%
	}
\label{ablation}
\end{table}

To quantitatively evaluate the impact of multi-scale strategy in our method, we present experimental results for various scale combinations, see Table \ref{ablation}. These experiments confirmed that the results only using the features of one scale have limited performance in distinguishing abnormal samples and normal samples. Instead, adopting a multi-scale approach brings the best performance to our model.
Furthermore, our observations indicate that resizing images to dimensions of 128$\times$128 and 192$\times$192 in a single-scale yields superior results compared to resizing to 256$\times$256 and 64$\times$64. This highlights the challenge of selecting the most appropriate image size for the task.
The combination of multi-scale images effectively alleviates this challenge, not only improving the overall performance of the model but also significantly enhancing its robustness.

\section{CONCLUSION}

We propose a new unsupervised anomaly detection model based on the principle of density estimation. Our method assumes that abnormal image features are distributed in low-density areas, while normal image features are concentrated in high-density areas. Utilizing AnoDODE, we perform density estimation of multi-scale image features, enabling efficient anomaly detection.
Combined with our model, we propose a tailored anomaly localization method that involves additional training of a decoder. This process produces image-level reconstructions and provides visual interpretability. Our method is robust provided by our multi-scale strategy and demonstrates state-of-the-art performance in anomaly detection on the BraTS2021 dataset.
In the future, we envision potential applications of our model in fields such as video detection. We also anticipate leveraging recent advances in diffusion ODE technology to further optimize our method.

\bibliographystyle{unsrtnat}
\bibliography{references}  %%% Uncomment this line and comment out the ``thebibliography'' section below to use the external .bib file (using bibtex) .

%\begin{thebibliography}{1}
%\bibitem{kour2014real}
 	%George Kour and Raid Saabne.
 	%\newblock Real-time segmentation of on-line handwritten arabic script.
 	%\newblock In {\em Frontiers in Handwriting Recognition (ICFHR), 2014 14th International Conference on}, pages 417--422. IEEE, 2014.



%\end{thebibliography}

\end{document}